\documentclass{article}

\usepackage{arxiv}

\usepackage[utf8]{inputenc} 
\usepackage[T1]{fontenc}    
\usepackage{hyperref}       
\usepackage{url}            
\usepackage{booktabs}       
\usepackage{amsfonts}       
\usepackage{nicefrac}       
\usepackage{microtype}      
\usepackage{lipsum}		
\usepackage{graphicx}
\usepackage{doi}

\title{High-fidelity social learning via shared episodic memories enhances collaborative foraging through mnemonic convergence}

\author{Ismael T. Freire\\
	Donders Institute for Brain, Cognition and Behaviour\\
	Radboud University\\
	Nijmegen, Netherlands \\
	\texttt{ismael.freire@donders.ru.nl} \\
    \And
	Paul Verschure \\
	Alicante Institute of Neuroscience, Department of Health Psychology\\
	Universidad Miguel Hernandez de Elche\\
	Elche, Spain\\
	\texttt{pverschure@umh.es} \\
}

\renewcommand{\shorttitle}{\textit{arXiv} Template}

\hypersetup{
pdftitle={High-fidelity social learning via shared episodic memories enhances collaborative foraging through mnemonic convergence},
pdfsubject={cs.MA},
pdfauthor={Ismael T. Freire, Paul Verschure},
pdfkeywords={Multi-agent Reinforcement learning, Episodic control, Social learning, Collective foraging},
}

\begin{document}
\maketitle
\begin{abstract} 
Social learning, a cornerstone of cultural evolution, enables individuals to acquire knowledge by observing and imitating others. At the heart of its efficacy lies episodic memory, which encodes specific behavioral sequences to facilitate learning and decision-making. This study explores the interrelation between episodic memory and social learning in collective foraging. Using Sequential Episodic Control (SEC) agents capable of sharing complete behavioral sequences stored in episodic memory, we investigate how variations in the frequency and fidelity of social learning influence collaborative foraging performance. Furthermore, we analyze the effects of social learning on the content and distribution of episodic memories across the group. High-fidelity social learning is shown to consistently enhance resource collection efficiency and distribution, with benefits sustained across memory lengths. In contrast, low-fidelity learning fails to outperform nonsocial learning, spreading diverse but ineffective mnemonic patterns. Novel analyses using mnemonic metrics reveal that high-fidelity social learning also fosters mnemonic group alignment and equitable resource distribution, while low-fidelity conditions increase mnemonic diversity without translating to performance gains. Additionally, we identify an optimal range for episodic memory length in this task, beyond which performance plateaus. These findings underscore the critical effects of social learning on mnemonic group alignment and distribution and highlight the potential of neurocomputational models to probe the cognitive mechanisms driving cultural evolution.
\end{abstract}


\section{Introduction}
Social learning can be defined as the process by which individuals acquire new knowledge, skills, attitudes, or behaviors by observing and imitating others within a social context \cite{whiten2000primate}. It involves the social transmission of information, where individuals learn from their interactions with others and the shared knowledge within their social group \cite{laland2004social}. At its core, social learning relies on the observation, imitation, and modeling of others' behavior \cite{bandura1977social}. Individuals observe the actions, behaviors, or outcomes of others and, based on these observations, learn and adopt similar behaviors or modify their existing behaviors \cite{munkenbeck1990social}. This learning process can occur through direct observation of others' actions, as well as through indirect forms such as verbal instructions, written materials, or media representations \cite{grusec1994social}. Social learning, therefore, encompasses various behaviors such as imitation, observational learning, and teaching, which contribute to cultural variation among human populations and may explain intraspecific variation in animal behavior \cite{whiten2007evolution, Gariepy2014}. 

One of the most telling examples of social learning is in the domain of foraging, seen in behaviors ranging from potato washing of Japanese monkeys to killer whale hunting methods, and even foraging traditions in wild birds such as the great tits \cite{kawamura1959process, aplin2015experimentally, black2023mammal, whiten2007evolution}. Such traditions are evidence of how learning and behavior are shaped by social influences. Although collective foraging is a widespread phenomenon observed in many animal species, humans stand out for their uniquely sophisticated collaborative skills, notably sharing knowledge and resources extensively within their social groups \cite{migliano2022origins, tomasello2011human}. But how does this uniquely human capacity to socially learn and collaborate actually emerged?




The interdependence hypothesis proposes that human collaborative and cognitive skills, which are essential for social learning, emerged to solve problems of social coordination in the context of mutualistic collaboration \cite{tomasello2012two}. More concretely, this hypothesis suggests that the unique forms of human cooperation and their underlying psychology originated in early humans when some changes in ecological conditions forced them to become obligate collaborative foragers \cite{tomasello_ultra-social_2014}. Under such conditions, human individuals became increasingly interdependent with one another, further motivating them to share more information and resources. 


Consequently, humans have developed a remarkable capacity to learn from others as well as to preserve and pass that knowledge to the next generation, contributing significantly to our success as a species in a process known as cumulative cultural evolution \cite{boyd_cultural_2011}. In particular, cumulative cultural evolution is a process that emerges when social learning allows for the preservation of information across generations, thereby enabling individual learning or fortunate transmission errors to refine it \cite{claidiere2010natural, montrey_evolution_2020}. The success of this process likely hinges more on the high fidelity of its transmission than on the efficiency of its refinement \cite{lewis2012transmission}. This is because as knowledge builds up, the failure of transmission necessitates the rediscovery or reinvention of a larger body of knowledge \cite{montrey_evolution_2020}. Still, understanding which are the precise cognitive mechanisms fueling this high-fidelity social learning is an active area of research \cite{thompson2022complex, mesoudi2016cultural, boyd1996culture, tennie2009ratcheting}.



Agent-based models have been traditionally employed to study collective foraging behavior and social learning dynamics \cite{couzin2005effective, conradt2005consensus, garg_individual_2022}. However, these models usually have limitations in capturing the complexity of cognitive processes, individual variation in behavior, and cognitive abilities among individuals \cite{baronchelli2018emergence, hawkins2019emergence, nisioti_social_2022}. 

In contrast, neuro-computational models of decision-making, such as reinforcement learning \cite{sutton2018reinforcement, mnih2015human}, are grounded in the physiology and psychology of human and animal cognition \cite{wyss2006model, Maffei2015, maffei2017perceptual, santos2021entorhinal}, and therefore allow for detailed simulations of individual and social learning processes. Moreover, these models usually provide a means to quantify properties like the diversity of experienced events, which are challenging to measure in human studies or agent-based models \cite{nisioti_social_2022}.

Building on the capabilities of neuro-computational models in the context of studying the cognitive mechanisms underlying social learning, the role of episodic memory becomes particularly relevant. Episodic control algorithms \cite{gershman2017reinforcement, Verschure2012}, which model memory- or instance-based learning processes supported by the mammalian hippocampus, facilitate the acquisition of robust behavioral strategies through storage and retrieval of past successful experiences \cite{Lengyel, santos2021epistemic}. These algorithms, which are more sample-efficient compared to state-of-the-art deep reinforcement learning models \cite{Blundell2016, zhu_episodic_2020}, make also direct use of stored memories to drive decision-making, making them particularly suitable for studying social learning among agents. 


In this paper, we investigate the role of episodic memory and social learning within the context of collective foraging by pursuing the following research questions: (Q1) How does sequential episodic memory affect high-fidelity social learning within the context of collective foraging? (Q2) What is the impact of the frequency and fidelity of social learning on the improvement of collective foraging? (Q3) How does the fidelity of social learning impact collective cognition? (Q4) Does a better distribution of collective knowledge among a community of foragers affect individual performance across agents? 


We hypothesize that (H1) the length of episodic memory will constrain the capacity of agents to learn complex behavioral sequences, therefore affecting their task performance; (H2) High-fidelity social learning will have a positive effect on collective foraging, magnified by the frequency of its occurrence; whereas low-fidelity will worsen performance the faster it happens (H3) Low-fidelity social learning will spread misleading information, leading to a more diverse but less aligned collective memory when compared with high-fidelity social learning; (H4) Only the homogeneous distribution of high-fidelity information will render a positive effect on the performance of individual agents, leading to a more equal distribution of rewards across the group.


The study comprises a series of experiments designed to test the above hypothesis regarding the role of episodic control in social learning and collective cognition during multi-agent foraging. We develop a collective foraging task in which agents need to gather food resources and bring them to the nest. To investigate the role of episodic memory in social learning (Q1), we implement Sequential Episodic Control (SEC) agents \cite{freire2023sec} endowed with the capacity to share complete behavioral sequences from its episodic memory and test them with varying memory lengths. The use of SEC in this context allows us to explore the potential of sequential representations in enhancing social learning, given its unique ability to store and communicate complete behavioral sequences. To study the role of fidelity and frequency of social learning (Q2), we manipulate (1) the transfer rate ($Tr$), which controls the number of trials in which agents can socially learn from each other, and (2) the transfer noise ($Tn$), that controls the accuracy at which sequential memories are transferred during social learning. This experimental setup allows us to explore how the communication of stored information can enhance collective problem-solving by looking at the trade-off between speed and accuracy during social learning. To elucidate the effects of social learning on collective memory (Q3), we evaluate the results of all experiments using several performance metrics, such as reward recollection and distribution, alongside mnemonic metrics which provide insights into the structure and dynamics of memory across agents \cite{coman_mnemonic_2016, momennejad_collective_2022}. Specifically, these include measures of individual mnemonic diversity (the variety of unique episodic memories retained by an individual agent), group mnemonic diversity (the diversity of episodic memories collectively shared across the group), and mnemonic group alignment (the degree of similarity between the long-term memories of agents within the group) \cite{nisioti_social_2022}. These mnemonic metrics help us understand how social learning influences the alignment, diversity, and distribution of memories at both individual and collective levels.

In conclusion, this research aims to elucidate the role of episodic memory in driving social learning during collaborative foraging and to explore the impact of length, speed, and fidelity of the social transmission of information. By examining these factors in the context of a biologically grounded cognitive model, SEC, we aim to provide a deeper understanding of the potential cognitive mechanisms underlying social learning in animals, including humans, and contribute to the growing body of knowledge on multi-agent reinforcement learning, social cognition, and cultural evolution. This work builds upon and significantly extends the preliminary results presented in our previous conference paper \cite{freire2023high}, by incorporating detailed analyses of mnemonic metrics and their role in shaping the dynamics and outcomes of social learning.

\section{Methods}

\subsection{Experimental setup} 
\begin{figure}[ht]
	\centering
	\includegraphics[width=\textwidth]{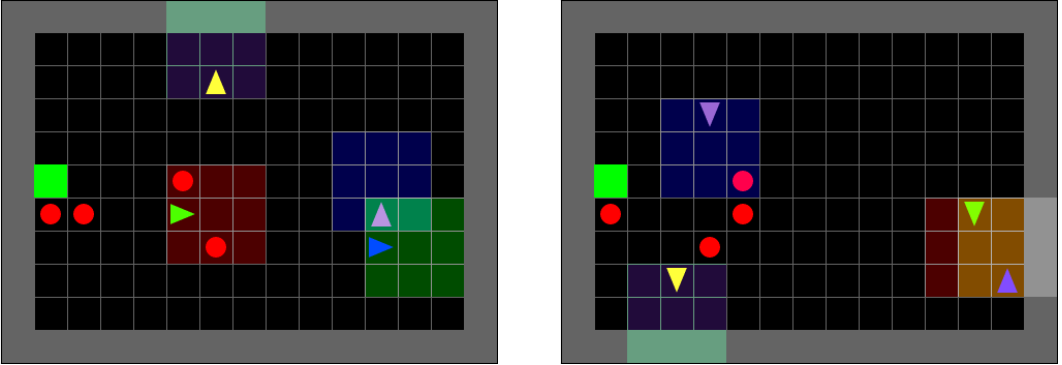}
	\caption{Collective foraging task modelled in a 2D grid-world environment. The environment contains four agents, four fruits, and a nest. Fruits are represented as red circles; the nest as a green square. The agents are represented as triangles of different colors (green, yellow, blue, and purple). The 3x3 colored area around each agent represents their field of view.}
	\label{fig:multigrid}
\end{figure}

We model the collaborative foraging task in a 2D discrete grid-world environment of size 11x15 containing four agents, four fruits, and a nest (see Figure \ref{fig:multigrid}). The goal of the agents is to pick the fruits and deposit them in the nest. For each fruit an agent deposits in the nest, it receives a positive reward of 1. Each agent can pass the fruit to or take it from another agent. An episode finishes when all the fruits have been collected or when it reaches 1000 timesteps. The agents play together for a total of 5000 episodes for each run.

Agents have partial observability, being able to only see a 3x3 section of the environment based on their position and orientation. Each grid cell is encoded with a tuple containing (1) the type of the object (such as wall, ball, agent, or goal), (2) the color of the object or other agent, (3) the type of the object that the other agent is carrying, (4) the color of the object that the other agent is carrying, (5) the direction of the other agent and (6) whether the other agent is actually one-self. 

The state $s_t$ of an agent is a 1d vector composed of the 3x3 grid tuples observed by the agent. Each action $a_t$ is encoded as a scalar number representing one of the following actions: turn left, turn right, move forward, pick up an object, and drop the object being carried.

We have implemented the collaborative foraging scenario using the multi-agent reinforcement learning modeling framework, Multigrid \cite{gym_multigrid2020}, a multi-agent extension of the Minigrid \cite{minigrid2018} library. These libraries contain a collection of discrete grid-world environments to conduct research on Reinforcement Learning. The environments follow the Gymnasium standard API and they are designed to be lightweight, fast, and easily customizable. 

\subsection{Sequential Episodic Control} 

This study employs the Sequential Episodic Control (SEC) algorithm to model foraging agents capable of storing past experienced events and learning from them \cite{freire2023sec}. The precise hyperparameters used in this experiment can be found in Appendix \ref{sec:appendix1:hyperparameters}. 

SEC is grounded in the Distributed Adaptive Control (DAC) theory, which considers the brain as a multi-layered control system \cite{Verschure2012,Verschure2014,verschure2016synthetic}. Previous cognitive models based on DAC have been instrumental in understanding how sensory-motor contingencies form and are exploited for behavioral control \cite{verschure2003environmentally, duff2011biologically, Maffei2015}. Building on this foundation, the Sequential Episodic Control (SEC) model extends DAC by incorporating key features of sequential learning, reflecting the key role of the hippocampus in episodic control \cite{freire2023sec}.

\begin{figure}[ht]
	\centering
	\includegraphics[width=\textwidth]{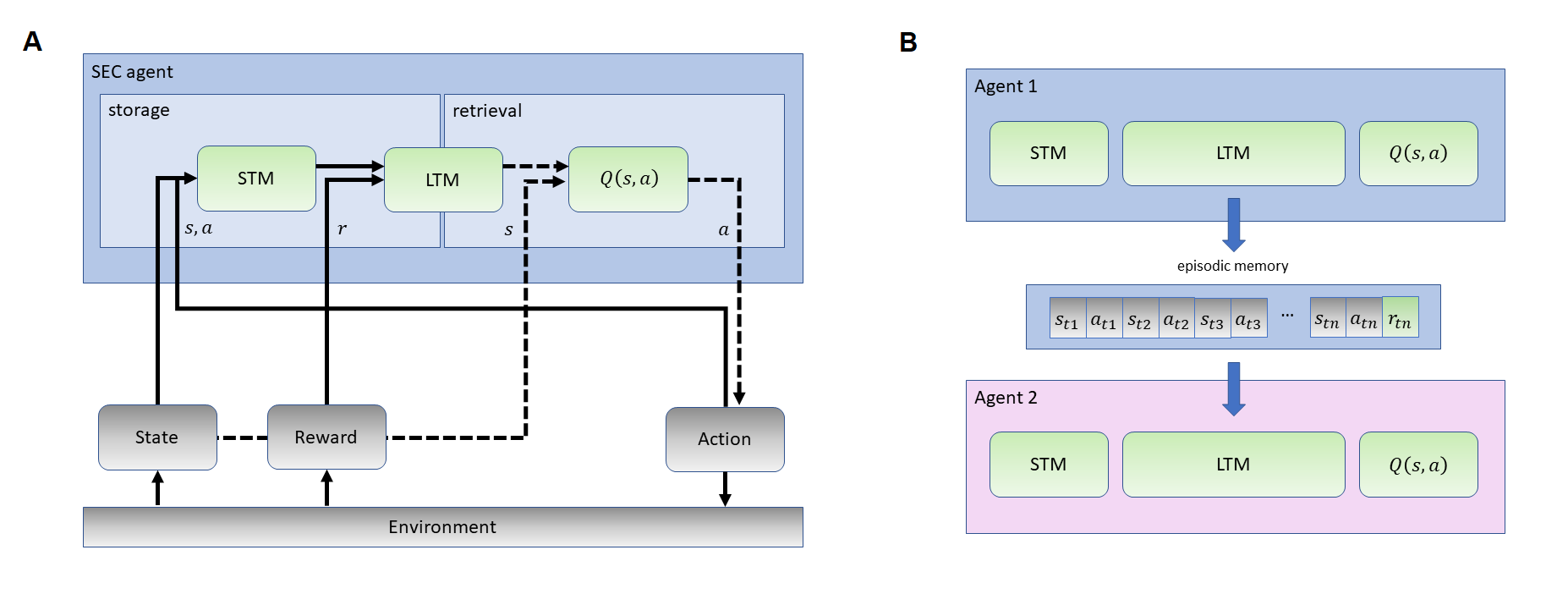}
	\caption{Panel A: Diagram of the SEC model for the 2D grid-world collective foraging task. SEC can be functionally divided into storage and retrieval phases. During the storage phase, the agent stores state-action ($s,a$) couplets into the short-term memory (STM) at each time step. When encountering a reward ($r$), the content of the STM buffer is transferred into the long-term memory (LTM). During the retrieval phase, the agent retrieves from its LTM the episodic memories more similar to the currently observed state. Based on this retrieved mnemonic information, the agent computes the state-action value function ($Q(s,a)$) for the current state and selects an action ($a$) based on the resulting probability distribution. Panel B: Social learning between two SEC agents. Agent 1 (blue) retrieves a copy of a complete episodic memory from its long-term memory (LTM) and transfers it to Agent 2 (pink), which stores the copy in its LTM.}
	\label{fig:SECmultiagent}
\end{figure}

SEC is a type of episodic control model designed to guide an agent's behavior based on previously rewarding state-action sequences, as opposed to classical episodic control algorithms that store state-action couplets in isolation \cite{Blundell2016, Pritzel2017, Lin2018}. In turn, SEC considers state-action pairs as integrated representational primitives and stores the complete sequence of state-action pairs leading to goal states, conserving their serial order, a key feature of the hippocampal function \cite{Buzsaki2018space}. This approach allows for the accumulation of knowledge and abilities that serve as the foundation for subsequent plans and behavior \cite{marcos2013hierarchical}. The SEC model comprises three key components: a short-term memory buffer (STM), a long-term episodic memory (LTM), and an action selection algorithm (see Figure \ref{fig:SECmultiagent}). 

As with standard episodic control cognitive models, SEC can be functionally divided into the storage and retrieval phases. During memory storage, SEC temporarily stores the most recent sequence of state-action pairs in the STM, while the LTM acts as a long-term repository for rewarding events. When the agent encounters a goal state, it associates its reward value with the current state-action sequence transiently maintained in the STM and consolidates the association between the STM and the reward in the LTM. 

In the memory retrieval phase, SEC selects memories from its long-term memory (LTM) that are relevant to its current state. This relevance is determined by an eligibility score, which is calculated based on how similar the current state is to the stored states in the LTM and how recently the memory was retrieved (for a detailed mathematical description of the algorithm, see \cite{freire2023sec}). Memories that surpass the similarity thresholds are then retrieved for action selection. The algorithm then calculates the value of each potential action based on the eligibility score of the selected memories and their associated rewards. The reward value is adjusted based on how close the memory is to the end of its sequence and is compared to the maximum reward of the selected memories. This process prioritizes actions that were taken in similar states and are closer to rewards in memory, reflecting their proximity in time and space in the real environment. Finally, the algorithm generates a probability distribution over the action space $Q(s,a)$, from which it samples the action to be performed. Initially, the algorithm explores actions randomly, but it becomes more selective as it accumulates more episodic memories.

\subsection{Social learning with Sequential Episodic Control} 
We consider a group of k = 4 Sequential Episodic Control (SEC) agents interacting together in the same environment and with the additional capacity to share past experienced events with each other. We model social learning in episodic control agents as the transmission of complete episodic memories between agents. In this context, the transmission of social information is local, decentralized, and based on what each agent perceives. Therefore, an agent can only share its memories with another agent present within its field of view. For instance, in the first scenario shown in Figure \ref{fig:multigrid}, only the blue agent would be capable of sharing its memories with the purple agent, as the latter falls within its field of view. In the second scenario depicted in Figure \ref{fig:multigrid}, the green and the purple agents can learn from each other reciprocally, as they both face one another.

Social learning between agents is governed by two key parameters: transfer rate $T_r$ and transfer noise $T_n$. Transfer rate determines the 'frequency' at which agents interact to share knowledge between them, while transfer noise determines the 'fidelity' of social learning. 

More concretely, $T_r$ controls the frequency of social learning between agents, in terms of episodes. A value of 1 means that agents can share information at every episode, while a value of 50 implies that they can only interact with each other every 50 episodes. Additionally, during each 'transfer episode', every agent has a small refractory period of 25 timesteps after each transmission to model the time invested in sharing the information in real time. 

Transfer noise, $T_n$, affects the fidelity of social learning by controlling the probability at which each element of the shared episodic memory is not correctly transferred between agents. Therefore, a value of $T_n = 0.1$ means there is a 10\% probability of information loss for each element composing the episodic memory. 

The episodic memories that are socially transmitted follow a process of 'prioritized experience sharing'\cite{nisioti_social_2022} (similar to prioritized experience replay \cite{schaul2015prioritized}), whereby memory sequences associated with higher reward values are shared more often. In other words, there is a prioritization in the retrieval of sequential memories proportional to the reward value associated with episodic memory. There is evidence of such a process of prioritization taking place during memory retrieval and replay in the rodent hippocampus \cite{foster2006reverse, olafsdottir2015hippocampal, mattar2018prioritized}.


\subsection{Mnemonic metrics} 
To analyze the impact of social learning through shared episodic memories, we apply the mnemonic metrics proposed in \cite{nisioti_social_2022}, with the addition of a new metric to assess memory distribution ($M_d$) between agents.

\begin{itemize}
    \label{methods:mnemonic_metrics}
    \item relative diversity $D_r$ is an agent-level metric that denotes the \textbf{number of unique episodic memories in an agent’s long-term memory}, relative to the total amount of memories acquired by the agent. In the results, we report an average $D_r$ over all agents in the group
	\item group diversity $D_g$ is a group-level metric that captures the \textbf{relative diversity of the aggregated group buffer}
	\item group alignment $A_g$ captures the similarity in terms of content between the long-term memories of agents belonging to the same group. To compute this we compute \textbf{the size of the common subset of episodic memories for each pair of agents} and, then, average over all these pairs, normalizing in [0,1].
    \item memory distribution $M_d$ is a novel group-level mnemonic metric that captures how evenly distributed are episodic memories across the group. 
\end{itemize}

\section{Results}

\subsection*{Memory length constraints agent performance and social learning, but not reward distribution} 

In the context of this study, the ability to store longer episodic memories translate into more complex behavioral sequences being shared and socially learned. 

\begin{figure}[!ht]
	\centering
	\includegraphics[width=\textwidth]{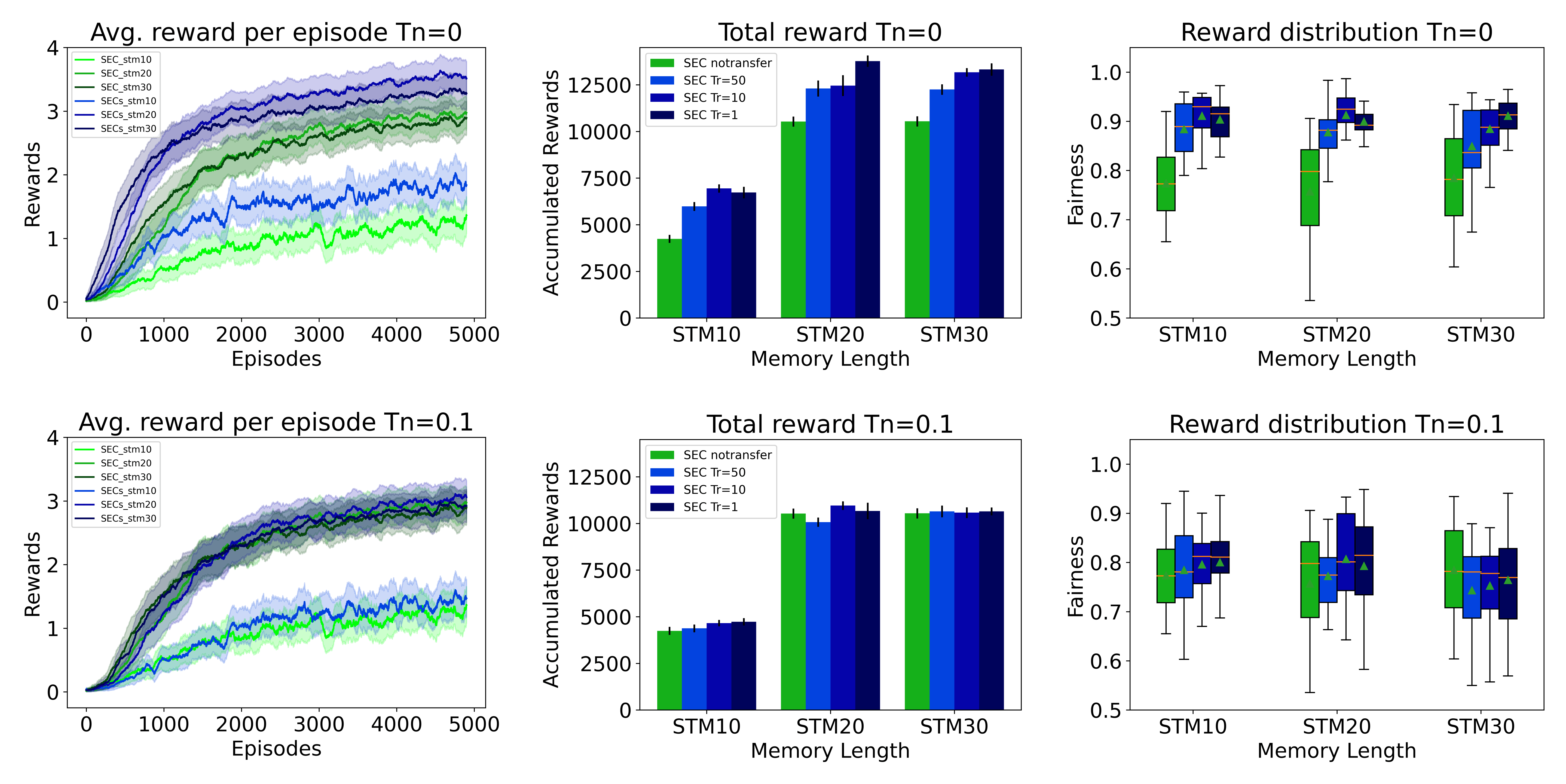}
	\caption{Performance results of Sequential Episodic Control agents across different social learning conditions. Top panels: results for high-fidelity social learning (Transfer Noise, Tn=0), Bottom panels: results for low-fidelity social learning (Tn=0.1). Left: average reward per episode. Center: total accumulated reward. Right: Ratio of reward distribution between agents. Colors represent the frequency of social interaction; green: no interaction (Tr=0), blue: infrequent interaction (every 50 episodes, Tr=50), royal blue: frequent interaction (every 10 episodes, Tr=10), continuous interaction (every episode, Tr=1). For clarity, only no interaction (Tr=0) and continuous interaction (Tr=1) results are shown on the left panels.}
	\label{fig:performance}
\end{figure}

Our study shows that memory capacity significantly impacts the performance of episodic control agents, a result consistent with previous research \cite{freire2021dacml, freire2023sec}. For instance, agents with their short-term memory constrained to 10 units (STM = 10) face a strong limitation of their capacity to obtain rewards, regardless of their ability to socially learn from their peers (see Figure \ref{fig:performance}).

However, more memory capacity does not always lead to better results. Interestingly, agents with larger short-term memory capacity (STM = 30) did not perform better than those with an average capacity (STM = 20). These results suggest that there is an optimal size for the episodic memories that might vary with the particular task and environmental configuration. 

Furthermore, the results show that memory capacity does not influence the distribution of rewards among agents (\ref{fig:performance}, right panels). As noted before, this effect is largely explained by the fidelity and frequency of social learning.

\subsection*{More frequent high-fidelity social learning improves resource collection and distribution, regardless of memory capacity} 

The performance results (see Figure \ref{fig:performance}, top panels) during high-fidelity social interactions show a general positive effect of the increase in the frequency of social transmission in terms of reward acquisition. This is true for the amount of rewards agents are able to collect each episode (Figure \ref{fig:performance}, top left) as well as for the total accumulated reward across episodes (Figure \ref{fig:performance}, top center). 

Regarding total accumulated reward, performance steadily increases along with the frequency of social interactions. This effect is consistent within each memory condition. In other words, more frequent high-fidelity social learning is beneficial for agents regardless of their memory capacity.

In addition, social learning frequency also affects how evenly distributed the rewards are among the population of agents (Figure \ref{fig:performance}, top right). An effect that also remains constant across all memory conditions.

\subsection*{Low-fidelity social learning offers no advantages over non-social learning} 
Regarding the performance metrics associated with low-fidelity social learning, the data suggests that such learning does not enhance the agents' ability to obtain rewards, both within individual episodes and overall (refer to Figure \ref{fig:performance}, bottom panels). Essentially, the outcomes from low-fidelity social learning mirror those of non-social learning agents with similar memory capacities, irrespective of how often social learning occurs. This starkly contrasts with the generally positive effect observed in reward acquisition as the frequency of high-fidelity social learning increased (see Figure \ref{fig:performance}, top panels). Furthermore, reward distribution is also influenced by low-fidelity social learning, aligning it with levels observed in agents under a non-social learning paradigm across all frequency conditions (Figure \ref{fig:performance}, bottom right).

\subsection*{Mnemonic diversity decreases during high-fidelity and frequent social learning, and increases with memory length} 

\begin{figure}[!ht]
	\centering
	\includegraphics[width=\textwidth]{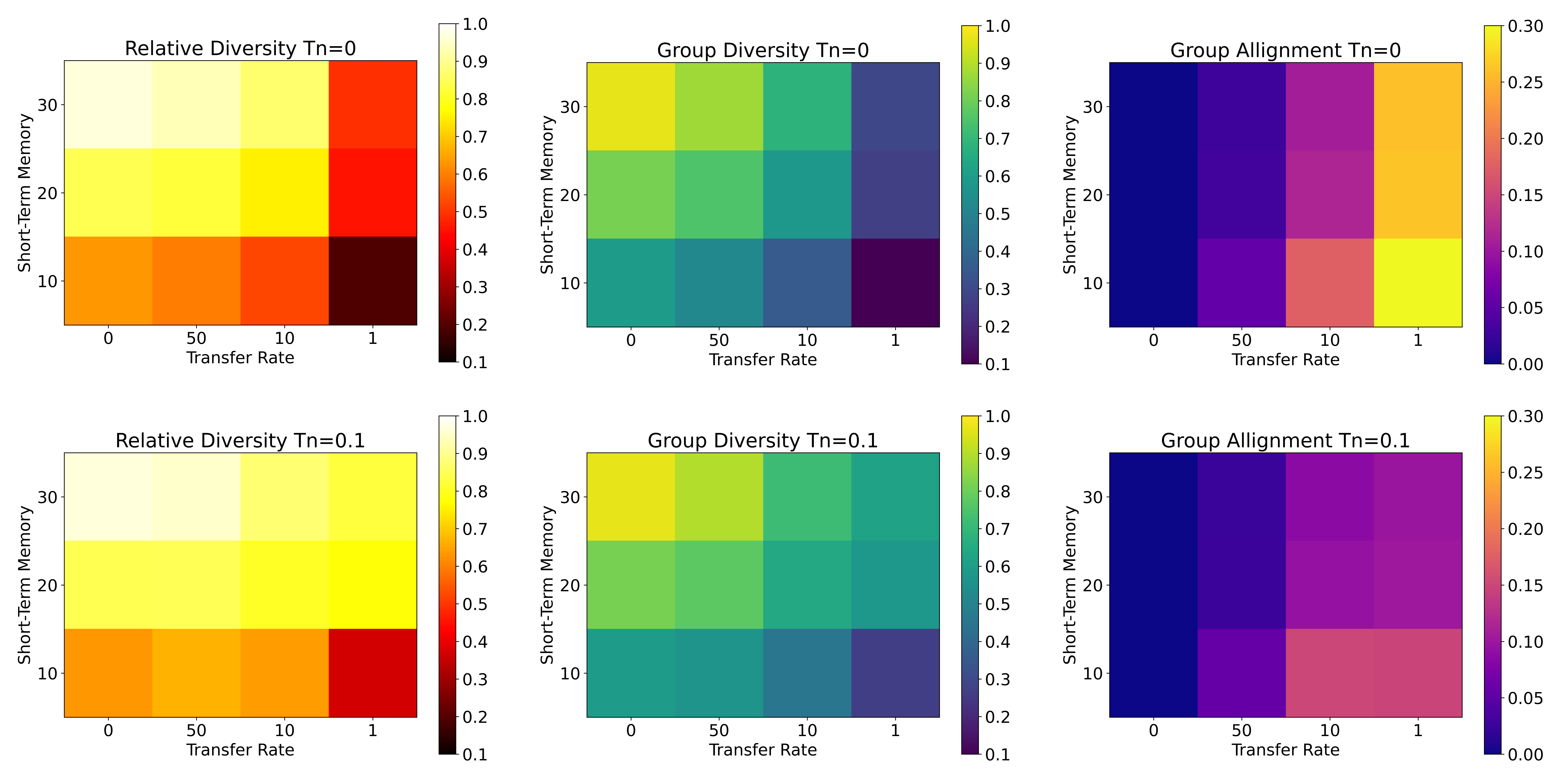}
	\caption{Mnemonic diversity results for both high-fidelity (top) and low-fidelity (bottom) social learning conditions. Left: average relative diversity (denotes the number of unique episodic memories in an agent’s long-term memory, relative to the total amount of memories acquired by the agent). Center: group diversity (a group-level metric that captures the relative diversity of the aggregated group buffer). Right: group alignment (captures the similarity in terms of content between the long-term memories of agents within a group).}
	\label{fig:mnemonic_diversity}
\end{figure}

Looking at the mnemonic diversity metrics during high-fidelity social learning (Figure \ref{fig:mnemonic_diversity}, top left, top center), we observe a double interaction effect caused by the agent's memory capacity and the frequency of social learning.

More concretely, longer memory buffers lead to higher relative and group diversity. We can observe that the baseline of both metrics is higher the longer the episodic memories, even for the non-social learning condition ($Tr$ = 0). Conversely, the effect of social learning frequency drives mnemonic diversity in the opposite direction: the more frequent the social transmission, the less mnemonic diversity the agents have. 

The most dramatic decrease in relative diversity (Figure \ref{fig:mnemonic_diversity} top left) can be precisely observed in the conditions of high-frequency social learning ($Tr$ = 1). This overall effect can also be observed when we look at mnemonic diversity at the group level (Figure \ref{fig:mnemonic_diversity} top center). These results indicate the accumulation of a greater number of similar rewards during high-frequency social learning, both at the group level and among individuals.


\subsection*{Mnemonic alignment increases during high-fidelity and frequent social learning, and decreases with memory length} 

Regarding mnemonic group alignment during high-fidelity social learning, we predictably observe a similar effect but in the opposite direction than with mnemonic diversity metrics, as it largely captures the opposite phenomenon at the group level: the degree at which the memories of the group have converged between agents. 

There is a key distinction to be made in this case. As we can observe in Figure \ref{fig:mnemonic_diversity} top right, the major driver of group alignment is the frequency of social learning, modulated to a small degree by the length of the episodic memories. The key role of social learning frequency in driving group alignment can be clearly seen by contrast with the non-social learning condition ($Tr$ = 0), where the results show group alignment is 0 for all memory conditions. In all other frequency conditions, the effect of alignment is amplified by the decreasing size of the memory buffer. Since there are fewer combinatorial possibilities for shorter episodic memories, the probabilities of alignment will increase faster with more frequent social learning between agents.

\subsection*{Low-fidelity social learning adversely impacts group alignment and simultaneously increases mnemonic diversity} 

As for the mnemonic diversity results in low-fidelity social learning, we observe a general increase in both relative diversity and group diversity across all frequency and memory conditions when compared one to one with the same conditions in the high-fidelity regime (Figure \ref{fig:mnemonic_diversity}, bottom panels). The pairwise difference between high and low fidelity is increasingly exacerbated when the frequency of social interactions is higher ($Tr$ = 1). The opposite effect holds for group alignment, where all low-fidelity conditions score lower when compared with the high-fidelity counterparts (Figure \ref{fig:mnemonic_diversity}, bottom right). 

The two mechanisms that help useful memories to propagate in high-fidelity conditions are the ones responsible for the propagation and amplification of noisy memories during low-fidelity social learning, namely: the frequency of social interactions and the length of the episodic memories. 

On the one hand, a higher frequency of low-fidelity social learning increases the chances of getting noisy copies of sequential memories, which is captured by the increase in diversity of high-frequency conditions when compared with the high-fidelity counterparts. On the other hand, the length of the short-term memory also affects the generation of more noisy memories, since longer episodic memories have more chances to suffer from noisy transmissions than smaller ones. These two facts, taken together, explain why the greater difference between high and low fidelity across all mnemonic metrics is seen in regimes of high frequency ($Tr$ = 1) and longer episodic memories (STM = 30).

\subsection*{Memory distribution strongly correlates with group alignment and reward distribution in high-fidelity social learning} 

\begin{figure}[!ht]
	\centering
	\includegraphics[width=\textwidth]{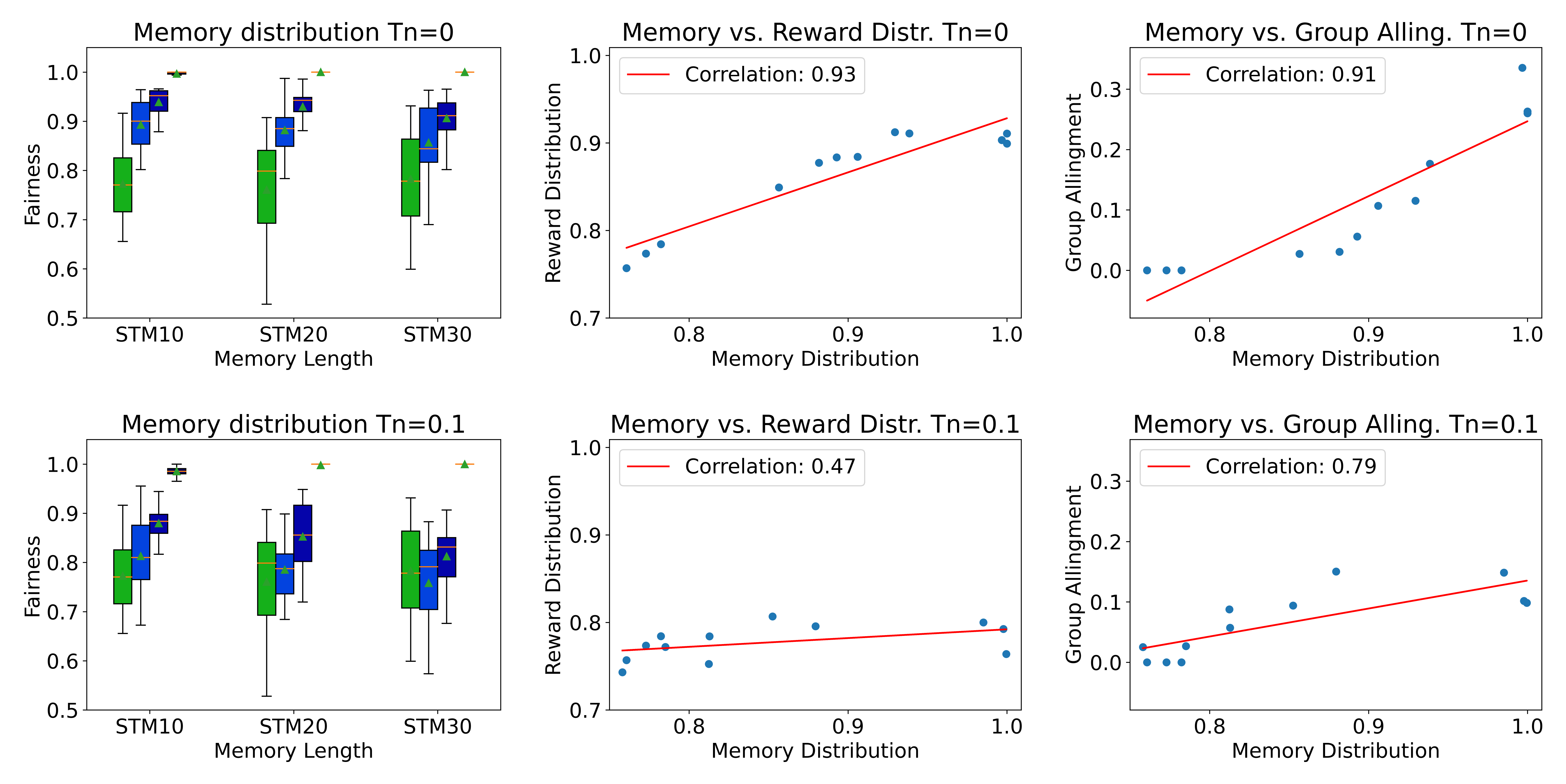}
	\caption{Mnemonic distribution strongly correlates with reward distribution and group alignment during high-fidelity social learning. Left panels: Mnemonic distribution results. Center: Correlations between memory distribution and reward distribution. Right: Correlations between memory distribution and group alignment. The top panels show results for high-fidelity social learning conditions. The bottom panels show results for low-fidelity social learning conditions.}
	\label{fig:memory_distribution}
\end{figure}

Memory distribution $M_d$ captures how well episodic memories are propagated across the agents. In figure  \ref{fig:memory_distribution}, left panels, we can observe a steady increase in memory distribution along with an increase in the frequency of social learning. This predictable increase in memory distribution with more frequent social learning is consistent across memory conditions, both during high-fidelity ($Tn$ = 0) and low-fidelity ($Tn$ = 0.1) social learning.

However, there is a key difference between high-fidelity and low-fidelity regimes regarding the quality of social learning that is clearly captured in two metrics: (1) behaviorally, by how well the distribution of memories across agents correlates with the distribution of rewards (see \ref{fig:memory_distribution}, mid panels); and (2) mnemonically, by how well memory distribution correlates with group alignment (see \ref{fig:memory_distribution}, right panels). 

As we can see, for high-fidelity social learning, there is a strong positive correlation between mnemonic and reward distributions ($r$ = 0.93), as well as between memory distribution and group alignment ($r$ = 0.91). As for low-fidelity social learning, the strength of both correlations is reduced; moderately for the correlation between memory distribution and group alignment ($r$ = 0.79), and strongly reduced for the correlation between memory and reward distributions ($r$ = 0.47).

In other words, this means that the distribution of memories between agents due to high-fidelity social learning drives mnemonic group alignment while also having a direct impact on the capacity of the agents to get rewards. On the other hand, the distribution of memories during low-fidelity social learning does have an impact on group alignment, but this mnemonic convergence does not translate into better reward distribution.

\section{Discussion}

In this paper, we have explored the role of social learning and episodic memory in collective foraging. We model social learning using sequential episodic control (SEC) agents capable of sharing complete behavioral sequences among them in the form of episodic memories stored in their long-term memory. We test them in a collective foraging task where the agents need to gather food resources and bring them to a nest. We perform a series of experiments in which we manipulate the length of episodic memories and the frequency and fidelity of social learning. 

Regarding our first research question (Q1), we found that the length of the stored episodic memory sequences indeed limits the capacity of agents to learn complex behavioral sequences, thereby affecting their task performance (H1). This was evident in our experiments where we manipulated the length of the episodic memories of Sequential Episodic Control (SEC) agents. Importantly, results have also shown that longer memory sequences do not necessarily yield better outcomes, indicating the existence of an optimal memory range for this task. Taken together with previous findings \cite{freire2023sec}, these results suggest that the optimal memory length for episodic control algorithms, such as SEC, is influenced, at least in part, by the specific demands of the task.


In response to our second research question (Q2), our results showed that high-fidelity social learning improves both total resource collection and resource distribution among agents compared to a non-social control condition (H2). This increase in performance was proportional to the frequency of social learning, and it disappeared in low-fidelity social learning conditions. However, when the fidelity of social learning was low, increasing the frequency did not improve performance. This is because agents were more likely to copy and propagate inaccurate or suboptimal behaviors. Therefore, although high-frequency, high-fidelity learning yields the best results, there is a nonlinear relationship between the speed of social learning (frequency) and the accuracy of the information being learned (fidelity).




Our third research question (Q3) concerned the effects of social learning on collective memory. We found that high-fidelity social learning increased mnemonic alignment (similar shared memories) and distribution between agents while decreasing mnemonic diversity (H3). In contrast, low-fidelity social learning reduced alignment and increased diversity by adding more noise into the memories of the agents. This result suggests that the fidelity of social learning may act as a gatekeeper, determining the extent to which collective knowledge influences exploration and exploitation at both individual and group levels. For example, in a foraging task, if all agents accurately copy a successful foraging strategy (high-fidelity learning), they will have similar memories and likely converge in the same areas, potentially depleting resources quickly. On the other hand, if agents learn less accurately (low-fidelity learning), they will have more diverse memories and might explore different areas, potentially leading to more sustainable use of resources but also increasing the risk of unsuccessful foraging. 

A recent study support this hypothesis, suggesting that communication noise and transmission errors in social learning can enhance collective problem-solving by maintaining diversity within a group \cite{boroomand2023superiority}. The diversity produced by low-fidelity social learning could be advantageous in scenarios where resources are distributed across distant patches, as a group with lower fidelity might explore more areas, potentially leading to better resource utilization. These findings suggest that the relationship between the fidelity of social learning and the exploration-exploitation role of collective knowledge might be more complex, and might also depend on the environmental constraints of each foraging context. An interesting avenue for future research would be to explore these interactions across a more diverse set of environments. For example, in foraging contexts with renewable but limited resources —akin to the Tragedy of the Commons scenario explored by \cite{perolat2017multi}— lower fidelity in social learning might mitigate resource overexploitation and promote a more sustainable strategy.

Beyond the potential use of social learning for exploration or exploitation, there is also the interplay between individual and social learning. A recent study by Garg and colleagues \cite{garg_individual_2022} examined how individuals in a group balance exploiting solutions discovered through social learning with independently exploring new, untested options. Their results show that while social learning is advantageous in rich and clustered environments, the benefits of exploiting social information are maximized when individuals also engage in high levels of individual exploration. They observe that the selective use of social information can mitigate the drawbacks of excessive social learning, particularly in larger groups and when individual exploration is limited. Further research is required in order to disentangle the complex relation of exploration-exploitation trade-offs between individual and social learning. 




Finally, in relation to our fourth research question (Q4), we observed that the distribution of high-fidelity memories among agents drove both mnemonic group alingment and the increase in reward distribution and overall performance (H4). On the other hand, the distribution of memories in low-fidelity social learning conditions did not have a positive effect on reward distribution among agents. This results underscores the balance between the quality of information being shared and its distribution among agents, and how the quality of information shared affects the performance of individual agents and the group as a whole.



Our findings provide meaningful insights into the relationships between social learning and episodic memory. Specifically, we identified nonlinear relationships between episodic memory length and task complexity (Q1-H1), fidelity and frequency of social learning (Q2-H2), and memory distribution and resource distribution (Q4-H4). Additionally, we propose that social learning fidelity mediates the exploration-exploitation dynamics of collective knowledge (Q3-H3).

Taken together, our results show that in collective foraging, there are complex relationships involving the fidelity of social learning, the diversity and alignment of collective memory, and the distribution of knowledge and resources among agents. In the collaborative foraging context studied in this paper, high-fidelity social learning leads to more aligned and less diverse memories within the group, and when these high-quality memories are widely and evenly distributed among agents, it leads to improved collective performance. However, low-fidelity social learning increases the diversity of memories within the group at the cost of reducing their alignment, and the wide distribution of these low-quality memories does not directly benefit collective performance in this context, although it might promote exploration. These results highlight the impact on collective behavior of both the quality and distribution of information in social learning. Additionally, it emphasizes the role of diversity and alignment of collective memory in adapting to dynamic environments and coordinating collective actions. While these effects are distinct, they are indeed interconnected, and it could be beneficial to conduct further studies involving the fidelity of social learning, the diversity and alignment of collective memory, and the distribution of memories and resources.

Despite our findings, this study faces several limitations that can be overcome in future research. First, regarding the resolution of the parameter search, it would be interesting to investigate intermediate levels of transfer noise $T_n$, between the current ones tested for high-fidelity social learning ($T_n$ = 0), and low-fidelity social learning ($T_n$ = 0.1). This will allow us to observe at which point the noisy transmission breaks down the positive effect of social learning, and whether that limit is different depending on the length of the shared episodic memories. Secondly, another aspect that can be investigated in more detail is the time course evolution of the mnemonic metrics across the episodes. This would give us greater temporal resolution to understand the dynamics and evolution of social learning and its relationship with the alignment of memories across a population.

Another important aspect that deserves more attention in the study of high-fidelity social learning in this context is how noisy information transmission differentially affects each of the components of a shared behavioral sequence. For instance, in this case, each episodic memory is composed of a series of states, actions, and a reward. The state, due to its larger size relative to the other components, is subject to more transmission errors, but each of them might have a lesser effect on the final result. It is likely that if a transmission error affects the reward component of the sequence, the impact on behavior will be much greater.
  
The complexity of shared information deserves further research in social learning. Recent literature suggests that the social transmission and preservation of complex algorithms is costly, necessitating selective social learning \cite{thompson2022complex}. Without selective social learning, subjects tend to converge on highly transmissible but lower-performance algorithms. Future modeling efforts could explore the potential trade-off between sequences that are simple and easily transmissible - and by default, might be less prone to transmission errors - and information that is more complex and challenging to share. There might be a trade-off between information complexity and transmission fidelity that could be worth investigating.

In our study of social learning using SEC agents, we deliberately chose to employ a 2D discrete grid-world environment as an abstraction for several reasons. First, the 2D grid-world abstraction allows us to create a controlled and simplified environment that focuses specifically on resource gathering and local social learning dynamics, without the complexities and confounding factors inherent in more realistic 3D environments. By reducing the dimensionality and scope of the environment, we can more easily isolate, parse, and investigate the key mechanisms and processes underlying social learning through the transmission of complete behavioral sequences. Furthermore, the use of a 2D grid-world abstraction provides a well-defined and easily interpretable framework for studying social interactions. The discrete nature of the grid-world environment allows us to precisely define the states, actions, and rewards, facilitating the analysis of agent behavior as well as the content of their memories. 

Although our study focuses on social learning within the context of a 2D grid world, we believe that the results obtained using SEC in this environment have broader implications and can generalize to other environments, as shown in previous studies with individual SEC agents \cite{freire2023sec}. By demonstrating the effectiveness of SEC agents in acquiring social learning skills in a controlled 2D setting, we provide evidence that these agents can potentially adapt and transfer their learned behaviors to more complex and continuous 3D environments. Moreover, the simplicity of the 2D grid-world abstraction allows us to establish a solid foundation and understanding of social learning mechanisms, which can serve as a basis for pursuing future research and applications in diverse environments. By building upon this foundation, researchers can explore the scalability of SEC to more realistic scenarios, considering factors such as high-dimensional and continuous state spaces \cite{de2020deep, amil2024discretization}, and complex spatial and combinatorial relationships between objects \cite{leibo2019autocurricula}.



Taken together, the use of episodic control models for studying the effects of social learning might be a promising avenue for further research with special relevance for cultural evolution, cognitive science, and the multi-agent reinforcement learning community. First, it allows us to study how particular properties at the cognitive level, such as the length of shared episodic memories, have an effect at the collective level, in terms of shared knowledge and collective performance. Second, it provides a state-of-the-art cognitive model that stores as episodic memories complete behavioral sequences of successful previous interactions, which can be useful to study the dynamics of cultural evolution across different generations of agents.

Our findings have significant implications for improving coordination in multi-agent systems through social learning mechanisms based on episodic control agents. The ability of these agents to share complete behavioral sequences from their episodic memory, particularly when this sharing is high-fidelity, can greatly enhance their collective performance. This is particularly relevant in situations that require the convergence of social conventions \cite{hawkins2019emergence} or in games of pure coordination \cite{freire2023modelingtom}.

For instance, in games like the Battle of the Sexes studied in \cite{Hawkins2016, freire2020conventions}, where the optimal strategy depends on the choices of other agents, the ability to accurately share and adopt successful strategies can lead to a faster convergence on a mutually beneficial equilibrium. High-fidelity social learning allows agents to accurately copy successful strategies, while the distribution of these high-quality memories ensures that all agents have access to this information. This can lead to more consistent and coordinated behavior across the group, reducing conflicts and improving overall performance.

The balance between memory diversity and alignment that we observed in our study can also be beneficial in these contexts. A certain level of memory diversity can allow for the exploration of different strategies and adaptability to changing environments or game dynamics. On the other hand, memory alignment can promote the adoption of successful strategies and the convergence of social conventions. These results highlight the potential of social learning mechanisms based on episodic control agents in enhancing coordination in multi-agent systems. Future research could further explore these mechanisms in different types of coordination games and real-world multi-agent scenarios, and investigate how to optimally balance the trade-offs we identified to maximize collective performance.

In conclusion, this study has highlighted the potential of episodic control models in examining the impacts of social learning on multi-agent collaborative tasks. This exploration paves the way for future research into the trade-offs between social and individual learning. Our findings aim to contribute to the fields of cultural evolution, cognitive science, and multi-agent reinforcement learning.

First, these models provide a unique perspective on the interplay between individual cognitive properties and collective outcomes. They allow us to investigate how specific individual cognitive traits, such as the extent of shared episodic memories, influence collective behaviors and shape shared knowledge, ultimately affecting collective performance \cite{nisioti_social_2022}. This offers a deeper comprehension of the role individual cognition plays in steering group dynamics and determining outcomes \cite{coman_mnemonic_2016, momennejad_collective_2022}.

Secondly, incorporating state-of-the-art cognitive models like the Sequential Episodic Control model \cite{freire2023sec} — which stores complete behavioral sequences from past successful interactions as episodic memories — provides fresh avenues to explore the dynamics of cultural evolution. Such models stand out as especially beneficial for simulating the transmission and evolution of cultural knowledge across agent generations, yielding insights into the underpinnings of cultural continuity and transformation.

In essence, our study reveals the potential of episodic control models in advancing our understanding of social learning and its implications for collective behavior. As we continue to explore this promising avenue, we look forward to uncovering new insights into the cognitive mechanisms underlying social learning, and their implications for cultural evolution and multi-agent systems.

\section*{Acknowledgments}
This study was supported by the Counterfactual Assessment and Valuation for Awareness Architecture (CAVAA) project of the EU Horizon-EIC program (ID: 101071178). 

\bibliographystyle{elsarticle-num}
\bibliography{bibliography}  

\appendix
\section{Sequential Episodic Control Hyperparameters} 
\label{sec:appendix1:hyperparameters}

SEC hyperparameter values for the collaborative foraging task. 

\begin{center}
\begin{tabular}{||c | c | c ||} 
 \hline
 Hyperparameter & Value & Description  \\ [0.5ex] 
 \hline\hline
 Episodes & 5000 & Episodes performed by each agent \\ 
 \hline
 LTM & 5000 & Long-term memory buffer length \\
 \hline
 State vector & 54 & Length of the 1d vector \\
 \hline
 $\beta_{i}$ & 0.1 & Increase of the bias term  \\
  \hline
 $\beta_{d}$ & 0.05 & Decay of the bias term  \\
 \hline
 $\tau$ & 0.9 & Decay factor  \\ 
 \hline
 $\theta_{prop}$ & 0.98 & Relative threshold   \\ 
 \hline
 $\theta_{abs}$ & 0.995 & Absolute threshold  \\ [1ex] 
 \hline
\end{tabular}
\end{center}

\end{document}